\documentclass[lettersize,journal]{IEEEtran}
\usepackage{amsmath,amsfonts}
\usepackage{algpseudocode}
\usepackage{array}
\usepackage[caption=false,font=normalsize,labelfont=sf,textfont=sf]{subfig}
\usepackage{textcomp}
\usepackage{stfloats}
\usepackage{url}
\usepackage{verbatim}
\usepackage{graphicx}
\usepackage{cite}
\usepackage{multirow}
\usepackage[hidelinks]{hyperref}
\hypersetup{
    colorlinks=false,   
    linktoc=all,        
    linkcolor=black,     
}
\usepackage{booktabs}
\usepackage{wrapfig}
\usepackage[linesnumbered,ruled,lined,boxed]{algorithm2e}
\usepackage[numbers,sort&compress]{natbib}
\usepackage{url}
\usepackage{makecell}
\usepackage{booktabs}

\begin{document}
\title{Weighted Joint Maximum Mean Discrepancy Enabled Multi-Source-Multi-Target Unsupervised Domain Adaptation Fault Diagnosis}

\author{Zixuan Wang,  Haoran Tang, Haibo Wang, Bo Qin, Mark D.\ Butala, \IEEEmembership{Member, IEEE}, Weiming Shen${^*}$, \IEEEmembership{Fellow, IEEE}, and Hongwei Wang${^*}$, \IEEEmembership{Member, IEEE}
\thanks{$^*$Corresponding authors}
\thanks{Zixuan Wang and Haibo Wang are with the College of Biomedical Engineering and Instrument Science in Zhejiang University, Hangzhou, 310013, China. (E-mail: \href{mailto:zixuanw.20@intl.zju.edu.cn}{zixuanw.20@intl.zju.edu.cn}).}
\thanks{Haoran Tang, Bo Qin, Mark D.\ Butala and Hongwei Wang are with Zhejiang University and the University of Illinois Urbana–Champaign Institute, Haining, 314400, China. (E-mail:  \href{mailto:haoran.21@intl.zju.edu.cn}{haoran.21@intl.zju.edu.cn}, \href{mailto:bo.20@intl.zju.edu.cn}{bo.20@intl.zju.edu.cn}, 
\href{mailto:markbutala@intl.zju.edu.cn}{markbutala@intl.zju.edu.cn}, \href{mailto:hongweiwang@intl.zju.edu.cn}{hongweiwang@intl.zju.edu.cn}).}
\thanks{Weiming Shen is with the State Key Laboratory of Digital Manufacturing Equipment and Technology, Huazhong University of Science and Technology, Wuhan 430074, China. (E-mail:  \href{mailto:wshen@ieee.org}{wshen@ieee.org}).}
}

\maketitle

\begin{abstract}
    Despite the remarkable results that can be achieved by data-driven intelligent fault diagnosis techniques, they presuppose the same distribution of training and test data as well as sufficient labeled data. Various operating states often exist in practical scenarios, leading to the problem of domain shift that hinders the effectiveness of fault diagnosis. While recent unsupervised domain adaptation methods enable cross-domain fault diagnosis, they struggle to effectively utilize information from multiple source domains and achieve effective diagnosis faults in multiple target domains simultaneously. In this paper, we introduce an original approach termed Weighted Joint Maximum Mean Discrepancy-enabled Multi-Source-Multi-Target Unsupervised Domain Adaptation (WJMMD-MDA), which represents a pioneering effort within the domain of fault diagnosis by facilitating domain adaptation in complex multi-source-multi-target settings for the first time. The proposed method efficiently captures pertinent information from multiple labeled source domains while effectuating domain alignment between these sources and the target domain via an improved weighted distance loss mechanism. Consequently, the model attains the acquisition of domain-invariant and discriminative features across multiple source and target domains, thereby enabling the implementation of cross-domain fault diagnosis. The performance of the proposed method is evaluated in comprehensive comparative experiments on three datasets, and the experimental results demonstrate the superiority of this method.
\end{abstract}


\section{Introduction}
\IEEEPARstart{W}{ith} the continuous development of big data technology in the industrial field, data-driven intelligent fault diagnosis techniques have been widely studied. Traditional machine learning methods, such as Random Forest (RF) \cite{cerrada2016fault}, LightGBM \cite{li2023lightgbm}, and Support Vector Machine (SVM) \cite{liu2016time}, have been widely used in the field of intelligent fault diagnosis. However, the dependence of machine learning-based fault diagnosis methods on domain knowledge and expertise leads to their lack of generalization ability. In recent years, deep learning \cite{7862893,kumar2021novel} has become a hot research trend in intelligent fault diagnosis because of its ability to extract features effectively with the aid of multilayer network structures. For example, \citet{wen2019new} proposed a hierarchical convolutional neural network that can predict the fault pattern and the fault severity. \citet{8097416} established a stacked sparse autoencoder-based deep neural network to diagnose the faults in rolling element bearings which reduces the need for prior knowledge and diagnostic expertise. \citet{9312649} decomposed the vibration signals of the wind turbine gearbox by wavelet packets and subsequently diagnosed the faults with a fast deep graph convolutional network.

However, deep learning-based fault diagnosis methods work only if there is a large amount of labeled data and if the training and test data have the same probability distribution. In practical situations, the collected data often come from various working states, resulting in different data distributions. Labeling data is both expensive and labor-intensive, making it extremely costly to label data for each working condition, which in turn constrains the practical application of deep learning-based methods.

Unsupervised domain adaptation (UDA) can cope with the above problem with its ability to learn domain-invariant features between labeled training data (source domain) and unlabeled test data (target domain). Unsupervised domain adaptation fault diagnosis methods effectively diagnose faults in the target domain by transferring the diagnostic knowledge from the labeled data of the source domain. For example, a domain adversarial graph convolutional network is proposed in \cite{9410617} to achieve unsupervised domain adaptation for mechanical fault diagnosis under variable working conditions. \citet{9247269} proposed a deep adversarial domain adaptation network to learn domain-invariant features for unsupervised domain adaptation. Researchers integrated expert knowledge with domain adaptation to realize unsupervised fault diagnosis in \cite{9612159}. 
However, existing unsupervised domain adaptation methods tend to be applicable only to single-source-single-target scenarios, while a typical drawback of such methods is that only insufficient representations can be extracted from a single supervised source domain. Fortunately, the latest studies \cite{xu2021ifds,xia2022moment,huang2021multisource} have begun to consider the extraction of more adequate features from multi-source domains for fault diagnosis in the target domain.

Despite the fact that the aforementioned studies have explored and obtained promising results for multi-source-single-target domain adaptation methods, they struggle to diagnose faults in multiple target domains. In the case of multi-target domains, it is not feasible to either train a diagnostic model for each target domain individually or to directly combine the unlabeled data from multiple target domains, as the former is costly while the latter would mix data with different distributions and different characteristics.

In this paper, we pioneered a multi-source-multi-target domain adaptation fault diagnosis method called WJMMD-MDA that can not only efficiently extract sufficient features from multiple labeled source domains, but also diagnose faults in multiple unlabeled target domains simultaneously. In the proposed method, data from multiple source and target domains are mapped into the feature space by the feature extractor with shared weights. Subsequently, feature alignment is implemented by the improved multi-source multi-target weighted joint maximum mean discrepancy (JMMD) distance for both source and target domains. By optimizing the weighted distance loss, domain-invariant and discriminative data features can be extracted for cross-domain fault diagnosis in the multi-source-multi-target scenario. 


The contribution of this paper is summarized as follows:
\begin{enumerate}
    \item We propose a novel multi-source-multi-target domain adaptation method for the first time in the field of fault diagnosis.
    \item We construct an improved weighted distance loss based on the maximum mean difference for multi-source-multi-target unsupervised domain adaptation.
    \item The excellent performance of the proposed method is demonstrated by comprehensive comparative experiments on three datasets, including the CWRU, PU, and PHM2009 datasets.
\end{enumerate}


The remainder of this paper is organized as follows. First, the related work is discussed in Section~II. Then, the detail of the proposed WJMMD-MDA is described in Section~III. The data used for the experiments and the comprehensive analysis of the results are shown in Section~IV. Finally, the conclusions of this paper are given in Section V.

\section{Related Work}
\subsection{Unsupervised domain adaptation}
In the past few years, there have been many methods proposed for domain adaptation, which aim to bridge the gap between the distribution of data in the source and target domains.
This is particularly useful when there is a lack of labeled data in the target domain, but abundant labeled data is available in a related source domain.
Theories and specific methods to achieve this objective have been established by prior researchers.
Existing works \cite{mansour2014robust, long2016unsupervised} provide theoretical proof for the possibility of transferring knowledge across domains when the training and test data features follow different distributions, whether single source domain or multiple source domains.
Metric discrepancy-based methods attempt to learn domain-invariant features by aligning the distributions of the source and target domains.
For example,  Maximum Mean Discrepancy (MMD) that maps the source and target domains to the reproducing kernel Hilbert space \cite{sejdinovic2013equivalence} is a popular distance metric used to estimate the similarity of the expectation between two distributions \cite{long2017deep, yan2017mind}. 
Moreover, there are amounts of methods with MMD \cite{long2018transferable, yan2019weighted} transfer the feature representation from the source domain to the target domain.
Since deep learning can achieve powerful feature representation, adversarial learning-based methods \cite{liu2019transferable, lee2019sliced} have also been widely applied in various fields, which involve training a model that can generalize well across domains by explicitly modeling the domain shift. 
Furthermore, Adversarial Domain Adaptation, using a discriminator to distinguish between source and target data, is a remarkable model-based approach.
Domain-Adversarial Training of Neural Networks (DANN) \cite{ganin2015unsupervised} and Adversarial Discriminative Domain Adaptation (ADDA) \cite{tzeng2017adversarial}, are two regular adversarial models for domain adaptation.
Specifically, DANN adds a domain classifier to the neural network and directly trains it to predict the domain label.
Conversely, ADDA learns a domain-invariant feature representation by training a generator network to transform source domain samples into target domain style and a discriminator network to distinguish between real and fake target domain samples.
To improve the generalization ability, \citet{cai2019unsupervised} propose a novel method with residual connections, which directly transforms the source features into the space of target features.
Generative adversarial networks (GANs) based methods \cite{kang2018deep, sankaranarayanan2018generate} have made significant advancements in generating samples that follow the distribution of target domains.
However, in real scenarios, there are often multiple source domains or multiple target domains, and domain adaptation methods with a single-source-single-target domain might not obtain the optimal solution.

\subsection{Multiple domains adaptation}
Nowadays, many researchers have made significant progress on the multiple domains adaptation problem. Numerous works have been done focusing on domain adaptation tasks from multiple source domains to a single target domain. For example, \citet{zhuNewMultipleSource2021} propose a multi-adversarial learning strategy for obtaining discriminative feature representations from multiple source domains. \citet{zhao2018adversarial} propose an adversarial network for multi-source domains that performs domain adaptation by optimizing task-adaptive generalization bounds. \citet{zhouProto2022} propose a prototype-based approach for multi-source domain adaptation, which aligns all source and target domains twice at the feature level to obtain better domain-invariant features and features closer to the class prototype, respectively.

The domain adaptation problem from a single source domain to multiple target domains has also received widespread attention. For example, an information-theoretic approach is proposed in \cite{ghoUnsuper2020a} for domain adaptation in the context of multiple unlabeled target domains and one labeled source domain. \citet{isoMultiTarget2021} propose a collaborative learning framework to achieve unsupervised multi-target domain adaptation. \citet{yangGraphAttention2022} propose a method for deep semantic information propagation in the context of multiple unlabeled target domains and a labeled source domain, where an attention mechanism can be used to optimize the relationship between multiple domain samples for better semantic transfer.
Distinguishing itself from prior research focusing on multi-domain adaptation, this paper pays more attention to the problem of multi-source-multi-target domain adaptation in the field of fault diagnosis. Given the prevalence of complex faults across diverse operational states in real industrial settings, this study addresses a gap in existing literature where limited efforts have been made to leverage multiple source domains for simultaneous fault diagnosis across multiple target domains.


\section{The proposed method}
\begin{figure*}[htbp]
\centering
\includegraphics[width=0.95\linewidth]{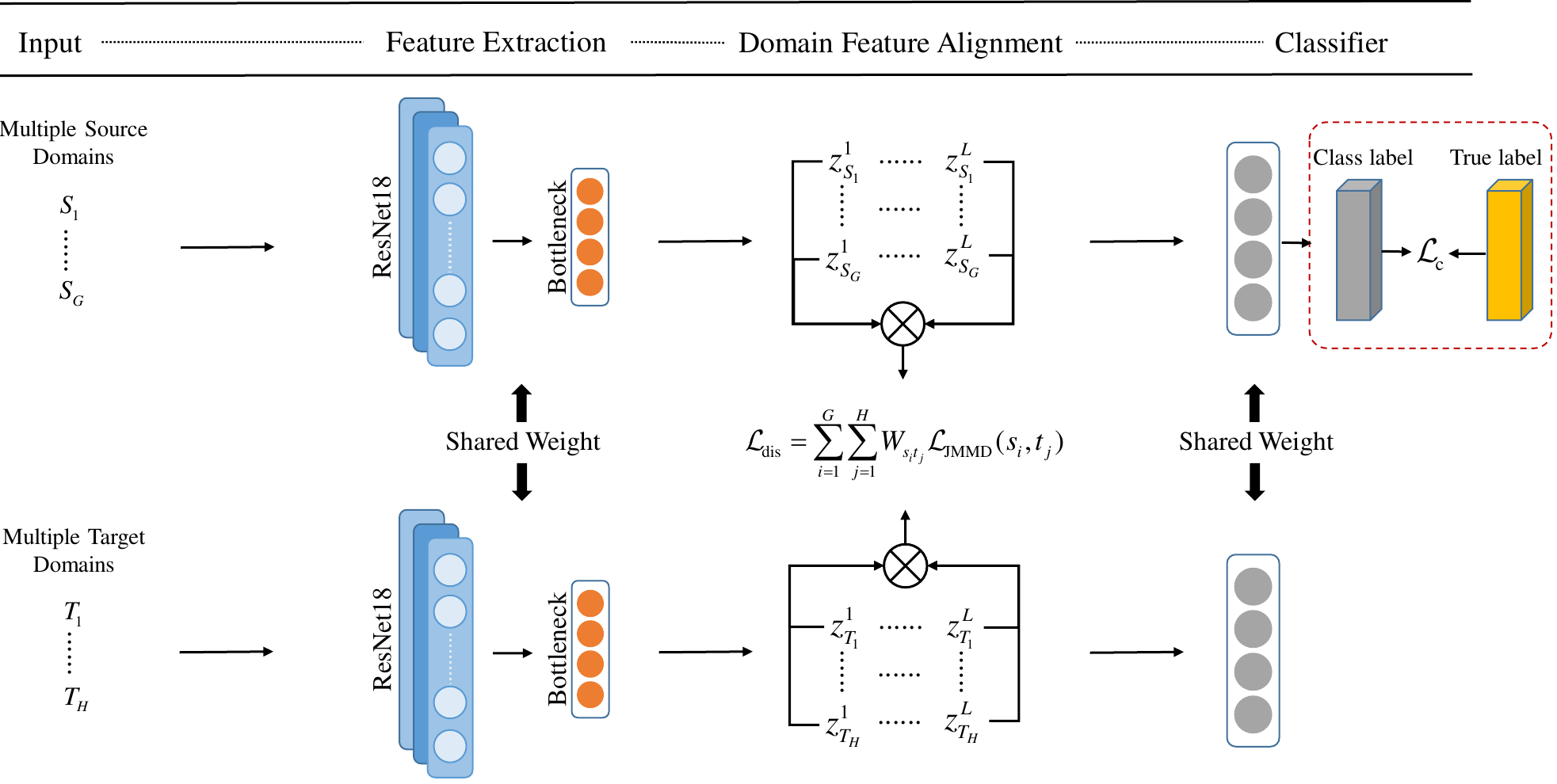}
\caption{The framework of the proposed weighted joint maximum mean discrepancy enabled multi-source-multi-target unsupervised domain adaptation fault diagnosis (WJMMD-MDA).}
\label{proposedmethod}
\end{figure*}

In the work, a novel multi-source-multi-target domain adaptation method is proposed, which utilizes an improved weighted distance loss based on the joint maximum mean discrepancy to capture the domain-invariant features between multiple source and target domains. To the best of our knowledge, this is the first time that a multi-source-multi-target domain adaptation method has been proposed in the field of fault diagnosis. 
The overview of the proposed method is shown in Figure \ref{proposedmethod}.

\subsection{Overview}
In our proposed method, signal data from multiple source and target domains are mapped into feature space by feature extractors with shared weights.
The feature extractor mainly includes two parts: convolution layers and bottleneck, where the bottleneck consists of a linear layer, a ReLU activation function and a dropout layer.
Subsequently, the feature alignment of the source domain and the target domain is achieved through the improved multi-source-multi-target weighted JMMD distance. 
By calculating the distance loss between different source domain and target domain data features and assigning weights, an improved multi-source multi-target weighted JMMD distance loss is obtained.
In addition, combined with the classification loss calculated by the classifier with shared weights, the final loss function is calculated.
Besides, the detailed structure of WJMMD-MDA is shown in Table \ref{networkstructure}.

\begin{table}[htb]
\tabcolsep=0.3cm
\renewcommand\arraystretch{1.5}
\centering
\caption{Detailed network structure of the proposed WJMMD-MDA.}
\label{networkstructure}
\begin{tabular}{@{}lccc@{}}
\toprule
Component & Layer & Filter & Output size                        \\ \midrule
\multirow{9}{*}[-8.5ex]{Feature extractor}&Input &\/ & 1$\times$512 \\
                                  &Conv\_layer\_1 &7$\times$7, 64, stride 2 & 64$\times$256  \\
                                  &Pooling\_1 &3$\times$3 max-pool & 64$\times$256  \\
                                  &Conv\_layer\_2 &
                                                    $\begin{bmatrix} 
                                                        3\times3, 64\\ 
                                                        3\times3, 64
                                                    \end{bmatrix} \times 2$
                                                    & 64$\times$128  \\
                                &Conv\_layer\_3 &
                                $\begin{bmatrix} 
                                    3\times3, 128\\ 
                                    3\times3, 128
                                \end{bmatrix} \times 2$
                                & 128$\times$64  \\
                                &Conv\_layer\_4 &
                                $\begin{bmatrix} 
                                    3\times3, 256\\ 
                                    3\times3, 256
                                \end{bmatrix} \times 2$
                                & 256$\times$32  \\
                                &Conv\_layer\_5 &
                                $\begin{bmatrix} 
                                    3\times3, 512\\ 
                                    3\times3, 512
                                \end{bmatrix} \times 2$
                                & 512$\times$16  \\
                                &Pooling\_2 &avg-pool\_view1 & 512   \\
                                &Bottleneck &512$\times$256 & 256  \\ \midrule
Classifier &Linear layer &256$\times$num\_classes &num\_classes\\
\bottomrule
\end{tabular}
\end{table}


\subsection{Objective function for multi-source-multi-target domain feature alignment}

The proposed method involves multiple labeled source domains $\left\{D_{s_1}, D_{s_2}, \cdots, D_{s_{\mathrm{G}}}\right\}$ and unlabeled target domains $\left\{D_{t_1}, D_{t_2}, \cdots, D_{t_{\mathrm{H}}}\right\}$, where $\mathrm{G}$ and $\mathrm{H}$ refer to the number of source domains and target domains, respectively.

In order to enable the proposed method with accurate classification of faults, we define the following classification loss function $\mathcal{L}_c^{s_i}$ for labeled samples $(x_j^{s_i}, y_j^{s_i})$ from source domain $D_{s_i}$:

\begin{equation}
\mathcal{L}_c^{s_i}=E_{\left(x_j^{s_i}, y_j^{s_i}\right) \sim D_{s_i}} l_{ce}\left(C\left(x_j^{s_i}\right), y_j^{s_i}\right)
\end{equation}
where $C(x_j^{s_i})$ refers to the predicted results, $E(\cdot)$ is the mathematical expectation, and $l_{ce}$ is the cross-entropy loss. The total classification loss function for multiple source domains is:

\begin{equation}
\mathcal{L}_c=\sum_{i=1}^G \mathcal{L}_c^{s_i}
\end{equation}

The MMD is commonly used to align the feature discrepancy  between the source domain and target domains, the distance loss function of MMD is defined in Eq. \eqref{mmdloss}, where $\phi(\cdot)$ represents the nonlinear mapping function to the reproducing kernel Hilbert space (RKHS), and $\Omega$ is the RKHS.


\begin{equation}
\label{mmdloss}
\begin{aligned}
\begin{split}
    \mathcal{L}_{\mathrm{MMD}} = & \left\|E_{x_i^s \sim D_s}\left[\phi\left(x_i^s\right)\right]\right. \\
    & \left. -E_{x_j^t \sim D_t}\left[\phi\left(x_j^t\right)\right]\right\|_{\Omega}^2 
\end{split}
\end{aligned}
\end{equation}



MMD has only considered the differences in features between domains but ignored the differences in labels between domains, therefore the Joint Maximum Mean Difference (JMMD) is proposed in \cite{long2017deep} on the basis of MMD. The loss function of JMMD is defined in Eq. \eqref{jmmdloss}, where $\otimes_{l=1}^{|L|} \phi^l\left(z_l^s\right)$ is the mapping of features in the tensor Hilbert space, while $z_l^s$ and $z_l^t$ stand for the activation of the $l$th layer of the source domain and target domain, respectively. 


\begin{equation}
\label{jmmdloss}
\begin{aligned}
\begin{split}
    \mathcal{L}_{\mathrm{JMMD}} = & \left\|E_{x_i^s \sim D_s}\left(\otimes_{l=1}^{|L|} \phi^l\left(z_l^s\right)\right)\right. \\
    & -\left.E_{x_j^t \sim D_t}\left(\otimes_{l=1}^{|L|} \phi^l\left(z_l^t\right)\right)\right\|_{\otimes_{l=1}^{\mid L l} \mathcal{\Omega}^l}^2
\end{split}
\end{aligned}
\end{equation}

In order to realize multi-source-multi-target domain adaptation, the proposed method calculates the JMMD distance $\mathcal{L}_{\mathrm{JMMD}}$ between two sets of source and target domains and performs a weighted summation of them as shown in Eq. \eqref{distanceloss}, where the weights are defined by Eq. \eqref{weight}. Specifically, when $\mathcal{L}_{\mathrm{JMMD}}$ between a pair of source and target domains is larger, a larger weight will be assigned, otherwise a smaller weight will be given.



\begin{equation}
\label{distanceloss}
\begin{aligned}
    \mathcal{L}_\text{dis} = \sum_{i=1}^G \sum_{j=1}^H W_{{s_i}{t_j}} \mathcal{L}_{\mathrm{JMMD}}({{s_i},{t_j}})
\end{aligned}
\end{equation}

\begin{equation}
\label{weight}
\begin{aligned}
 W_{{s_i}{t_j}}=\frac{e^{\mathcal{L}_{\mathrm{JMMD}}({{s_i},{t_j}})}}{\sum_s\sum_t e^{\mathcal{L}_{\mathrm{JMMD}}(s,t)}}
 \end{aligned}
\end{equation}

The overall loss function is shown below, where $\lambda$ is a trade-off parameter. By optimizing the overall loss function in Eq. \eqref{overall_loss}, feature alignment between multiple source domains and multiple target domains can be achieved. Sufficient diagnostic information can be effectively transferred to the unlabeled target domains, enabling domain adaptation fault diagnosis under multi-source-multi-target scenarios.

\begin{equation}
\label{overall_loss}
\begin{aligned}
    \mathcal{L} = \lambda \mathcal{L_\text{c}} + (1-\lambda) \mathcal{L_\text{dis}}
\end{aligned}
\end{equation}

\subsection{Training procedure and parameter update}
The overall training process is shown in Algorithm \ref{msmtalgorithm}. 
The input datasets consist of labeled samples from multiple source domains $\left\{\left(x_{s_i}, y_{s_i}\right)\right\}_{i=1}^{\mathrm{G}}$ and unlabeled samples from multiple target domains $\left\{\left(x_{t_j}\right)\right\}_{j=1}^{\mathrm{H}}$.
We will first initialize the model parameters, including the parameters of the feature extractor $\theta_f$ and classifier $\theta_y$, as well as hyper-parameters, and perform batch processing on the input datasets. 
Each batch is selected for model training, and the gradient descent method is used to update the model parameters $\left\{\theta_f, \theta_y\right\}$ to minimize the overall loss, including classification loss $\mathcal{L_\text{c}}$ and distance loss $\mathcal{L_\text{dis}}$, until the model is converged.

\begin{algorithm}[htb]
\label{msmtalgorithm}
\caption{Weighted JMMD Enabled Multi-source-multi-target Domain Adaptation Fault Diagnosis}
\KwIn{Labeled samples from multiple source domains $\left\{\left(x_{s_i}, y_{s_i}\right)\right\}_{i=1}^{\mathrm{G}}$ and unlabeled samples from multiple target domains $\left\{\left(x_{t_j}\right)\right\}_{j=1}^{\mathrm{H}}$, learning rate $\eta$, batch\_size $bs$, epochs $n$, model $f$ and $y$ with parameters $\theta_f$ and $\theta_y$}
\KwOut{Trained model with parameters $\theta^{*}$}
Initialize $\theta_f$ and $\theta_y$; \\
\Repeat{\text{Convergence}}{
    Sample a mini-batch from each of the source/target domain datasets;\\
    Update $\left\{\theta_f, \theta_y\right\}$ by minimizing the overall loss function $\mathcal{L}$ in Eq. \eqref{overall_loss} through gradient descent $\theta_f=\theta_f-\frac{\partial \mathcal{L}}{\partial \theta_f}$ and $\theta_y=\theta_y-\frac{\partial \mathcal{L}}{\partial \theta_y}$;\\
    }
\end{algorithm}

\section{Experiment} 
\subsection{Datasets}

\begin{table}[htb]
\centering
\caption{10 bearing conditions of CWRU dataset.}
\label{CWRUfault}
\resizebox{0.30\textwidth}{!}{ 
\begin{tabular}{@{}clc@{}}
\toprule
Class label & Bearing condition & Fault size (mils) \\ \midrule
0   & Normal bearing & 0   \\
1   & Inner ring fault & 7 \\
2   & Ball fault  & 7   \\
3   & Outer ring fault & 7 \\
4   & Inner ring fault &  14 \\
5   & Ball fault& 14  \\
6   & Outer ring fault  & 14 \\
7   & Inner ring fault &  21 \\
8   & Ball fault  & 21 \\
9  & Outer ring fault & 21 \\
\bottomrule
\end{tabular}}
\end{table}

\begin{table}[htb]
\centering
\caption{Domain adaptation tasks of CWRU dataset.}
\label{domainCWRU}
\resizebox{0.30\textwidth}{!}{ 
\begin{tabular}{@{}lcccc@{}}
\toprule
Working condition & $\mathrm{C}_0$ & $\mathrm{C}_1$ & $\mathrm{C}_2$ & $\mathrm{C}_3$\\ \midrule
Motor rotational speed (rpm)   & 1797 & 1772 & 1750 & 1730  \\
\bottomrule
\end{tabular}}
\end{table}

\begin{table}[htb]
\centering
\caption{Domain adaptation tasks of PU dataset.}
\label{domainPU}
\resizebox{0.30\textwidth}{!}{ 
\begin{tabular}{@{}lcccc@{}}
\toprule
Working condition & $\mathrm{P}_0$ & $\mathrm{P}_1$ & $\mathrm{P}_2$ & $\mathrm{P}_3$\\ \midrule
Rotating speed (rpm)   & 1500 & 900 & 1500 & 1500  \\
Load torque (Nm)   & 0.7 & 0.7 & 0.1 & 0.7  \\
Radial force (N)   & 1000 & 1000 & 1000 & 400  \\
\bottomrule
\end{tabular}}
\end{table}

\begin{table}[htb]
\centering
\caption{Domain adaptation tasks of PHM2009 dataset.}
\label{domainPHM}
\resizebox{0.30\textwidth}{!}{ 
\begin{tabular}{@{}lcccc@{}}
\toprule
Working condition & $\mathrm{M}_0$ & $\mathrm{M}_1$ & $\mathrm{M}_2$ & $\mathrm{M}_3$\\ \midrule
Shaft speed (Hz)   & 30 & 35 & 40 & 45  \\
\bottomrule
\end{tabular}}
\end{table}

Three datasets are used in the following experiments. A detailed description of them is shown below.
\begin{enumerate}
    \item Case Western Reserve University (CWRU) dataset: The CWRU dataset is provided by the Case Western Reserve University Bearing Data Center \cite{loparo2012case} and has been widely used in the field of fault diagnosis. Referring to the setting in \cite{zhao2021applications}, this paper uses the driving end bearing fault data with a sampling frequency of 12kHz, which includes one normal signal and nine fault signals The details of the 10 categories of signals are shown in Table \ref{CWRUfault}, which include three positions of inner fault (IF), outer fault (OF), and ball fault (BF), and three different sizes of fault signals are selected for each position. Domain adaptation experiments were carried out at four different operating speeds, as shown in Table \ref{domainCWRU}.
    \item Paderborn University (PU) Dataset: The PU dataset is a bearing dataset collected from the test rig of Paderborn University Bearing Data Center with a sampling frequency of 64 kHz, including artificially induced damages and real damages \cite{lessmeier2016condition}. According to \cite{zhao2021applications}, a total of 13 real bearing faults were selected to carry out domain adaptation fault diagnosis experiments under 4 different working conditions. The bearing code of the 13 faults are KA04, KA15, KA16, KA22, KA30, KB23, KB24, KB27, KI14, KI16, KI17, KI18, and KI21. The detailed description of  4 working conditions is shown in Table \ref{domainPU}.
    \item PHM Data Challenge on 2009 (PHM2009) Dataset: The PHM2009 dataset is a generic industrial gearbox dataset provided by the PHM Data Challenge competition, including spur and helical gears \cite{challenge2009available}. Based on the settings in \cite{zhao2021applications}, we selected the helical gear dataset (it has 1 healthy condition and 5 faults) with four shaft speeds under high loads, as the four working conditions. The details of the four working conditions are shown in Table \ref{domainPHM}.
\end{enumerate}

\subsection{Experimental setup}
We compare the proposed method with state-of-the-art techniques: JMMD \cite{long2017deep}, MK-MMD \cite{gretton2012optimal}, CORAL \cite{sun2016deep}, DANN \cite{ganin2016domain}, CDAN \cite{long2018conditional} and SDAFDN \cite{lu2022self}. A brief introduction of the baselines is shown below. 

\begin{enumerate}
    \item[(a)] JMMD: It achieves domain alignment by minimizing the joint distribution distance between the source and target domain.
    \item[(b)] MK-MMD: It achieves domain alignment by optimizing the weighted sum of maximum mean discrepancy (MMD) based on multiple kernel functions.
    \item[(c)] CORAL: It achieves domain adaptation by aligning the second-order statistics of the source and target domain distributions.
    \item[(d)] DANN: It achieves domain adaptation by adversarial training of source and target domains based on domain discriminators and feature extractors.  
    \item[(e)] CDAN: It realizes multimodal distribution alignment by adding multilinear conditioning to DANN.
    \item[(f)] SDAFDN: It achieves domain adaptation based on feature mapping and domain adversarial training, as well as down-sampling and interaction networks for time dependency. For SDAFDN, we directly report the results from the original paper \cite{lu2022self}, so the comparison experiments on the CWRU dataset do not have results from this method.
\end{enumerate}

For the fairness of the experiments, the feature extractors and classifiers involved in the comparison methods are the same as those in the proposed method, except for SDAFDN, whose results are reported directly.

In our experiments, we use Z-score normalization for data preprocessing. 
The data sample length was set to 1024 and a Fast Fourier Transform (FFT)  was applied to the data sample. For each working condition, 80\% of the samples were divided as the training set and 20\% of the samples were divided as the test set.  

\subsection{Comparion with other methods}

\begin{table*}[ht] 
    \centering
    \caption{Transfer learning diagnosis results on CWRU dataset}
    \label{CWRU}
    \resizebox{\textwidth}{!}{ 
    \begin{tabular}{|l|c|c|c|c|c|c|c|c|c|c|c|c|c|}
    \hline
        \textbf{Source Domain} & \multicolumn{2}{c|}{$\mathrm{C}_0\quad+\quad\mathrm{C}_1$} & \multicolumn{2}{c|}{$\mathrm{C}_2\quad+\quad\mathrm{C}_0$} & \multicolumn{2}{c|}{$\mathrm{C}_3\quad+\quad\mathrm{C}_0$} & \multicolumn{2}{c|}{$\mathrm{C}_1\quad+\quad\mathrm{C}_2$} & \multicolumn{2}{c|}{$\mathrm{C}_3\quad+\quad\mathrm{C}_1$} & \multicolumn{2}{c|}{$\mathrm{C}_3\quad+\quad\mathrm{C}_2$} &  \multirow{2}{*}{\textbf{Average}} \\ \cline{1-13}
        \textbf{Target Domain} & \multicolumn{2}{c|}{$\mathrm{C}_2 \qquad \quad \mathrm{C}_3$} & \multicolumn{2}{c|}{$\mathrm{C}_1 \qquad \quad \mathrm{C}_3$} & \multicolumn{2}{c|}{$\mathrm{C}_2 \qquad \quad \mathrm{C}_1$} & \multicolumn{2}{c|}{$\mathrm{C}_0 \qquad \quad \mathrm{C}_3$} & \multicolumn{2}{c|}{$\mathrm{C}_0 \qquad \quad \mathrm{C}_2$} & \multicolumn{2}{c|}{$\mathrm{C}_1 \qquad \quad \mathrm{C}_0$} &  \\ \hline
        JMMD & 0.8539  & 0.9968  & 0.8321  & 1.0000  & 0.9805  & 0.9903  & 0.9195  & 1.0000  & 0.8774  & 1.0000  & 0.9091  & 0.8467  & 0.9339  \\ \hline
        MK-MMD & 1.0000  & 1.0000  & 0.9968  & 1.0000  & 1.0000  & 1.0000  & 0.8812  & 0.9903  & 0.8659  & 1.0000  & 0.9773  & 0.7586  & 0.9558  \\ \hline
        CORAL & 1.0000  & 0.9612  & 0.9740  & 0.8608  & 1.0000  & 0.9935  & 0.9579  & 1.0000  & 0.9655  & 1.0000  & 0.8994  & 0.8352  & 0.9540  \\ \hline
        DANN & 1.0000  & 0.9935  & 0.9675  & 1.0000  & 0.9935  & 1.0000  & 0.9540  & 0.9676  & 0.9349  & 1.0000  & 0.9643  & 0.9770  & 0.9794  \\ \hline
        CDAN & 0.9123  & 0.9223  & 0.9545  & 0.8576  & 1.0000  & 1.0000  & 0.9617  & 1.0000  & 0.8352  & 1.0000  & 0.9318  & 0.9425  & 0.9432  \\ \hline
        \textbf{Proposed method} & 1.0000  & 1.0000  & 0.9935  & 0.9644  & 1.0000  & 1.0000  & 1.0000  & 1.0000  & 0.9962  & 1.0000  & 1.0000  & 0.9923  & \textbf{0.9955}  \\ \hline
    \end{tabular}}
\end{table*}
For single-source-single-target domain adaptation tasks, each column represents the corresponding source and target domains. For example, the first column in Table \ref{CWRU} represents the domain adaptation task from source domain 0 to target domain 2. For multiple-source-multi-target domain adaptation tasks, the combined columns represent the corresponding multiple source domains and multiple target domains. For example, the first group column $\mathrm{C}_0$/$\mathrm{C}_1$ $\rightarrow$ $\mathrm{C}_2$/$\mathrm{C}_3$ in Table \ref{CWRU} represents the domain adaptation tasks that transfer from source domains $\mathrm{C}_0$ and $\mathrm{C}_1$ to target domains $\mathrm{C}_2$ and $\mathrm{C}_3$.

Tables \ref{CWRU}, \ref{PU}, and \ref{PHM} show the domain adaptation fault diagnosis results of the proposed method as well as the comparison methods on CWRU, PU, and PHM2009 datasets, respectively. As can be seen from the tables, the results obtained by the proposed method have a great advantage over the results obtained by the compared methods on all datasets. In particular, for the most challenging PHM2009 dataset, the average accuracy obtained by the proposed method is 12.86\% higher than the best result obtained by the compared methods, as shown in Table \ref{PHM}. It demonstrates that the proposed method fully utilizes the information from multiple source domains and effectively performs domain adaptation for multiple target domains compared with the single-source-single-target methods. The excellent domain adaptation fault diagnosis results shown in Tables \ref{CWRU}, \ref{PU}, and \ref{PHM} demonstrate the superior performance of the proposed multi-source-multi-target approach, and it is noteworthy that the proposed approach is the first application of the concept of multi-source-multi-target domain adaptation in the field of fault diagnosis.
\begin{table*}[ht] 
    \centering
    \caption{Transfer learning diagnosis results on PU dataset}
    \label{PU}
    \resizebox{\textwidth}{!}{ 
    \begin{tabular}{|l|c|c|c|c|c|c|c|c|c|c|c|c|c|}
    \hline
        \textbf{Source Domain} & \multicolumn{2}{c|}{$\mathrm{P}_0\quad+\quad\mathrm{P}_1$} & \multicolumn{2}{c|}{$\mathrm{P}_2\quad+\quad\mathrm{P}_0$} & \multicolumn{2}{c|}{$\mathrm{P}_3\quad+\quad\mathrm{P}_0$} & \multicolumn{2}{c|}{$\mathrm{P}_1\quad+\quad\mathrm{P}_2$} & \multicolumn{2}{c|}{$\mathrm{P}_3\quad+\quad\mathrm{P}_1$} & \multicolumn{2}{c|}{$\mathrm{P}_3\quad+\quad\mathrm{P}_2$} &  \multirow{2}{*}{\textbf{Average}} \\ \cline{1-13}
        \textbf{Target Domain} & \multicolumn{2}{c|}{$\mathrm{P}_2 \qquad \quad \mathrm{P}_3$} & \multicolumn{2}{c|}{$\mathrm{P}_1 \qquad \quad \mathrm{P}_3$} & \multicolumn{2}{c|}{$\mathrm{P}_2 \qquad \quad \mathrm{P}_1$} & \multicolumn{2}{c|}{$\mathrm{P}_0 \qquad \quad \mathrm{P}_3$} & \multicolumn{2}{c|}{$\mathrm{P}_0 \qquad \quad \mathrm{P}_2$} & \multicolumn{2}{c|}{$\mathrm{P}_1 \qquad \quad \mathrm{P}_0$} &  \\ \hline
        JMMD & 0.9511  & 0.4387  & 0.6350  & 0.8805  & 0.9023  & 0.4724  & 0.5223  & 0.8245  & 0.7450  & 0.6519  & 0.4018  & 0.9032  & 0.6941  \\ \hline
        MK-MMD & 0.9527  & 0.4039  & 0.6043  & 0.7988  & 0.8122  & 0.5245  & 0.5868  & 0.8578  & 0.7588  & 0.5863  & 0.3880  & 0.9324  & 0.6839  \\ \hline
        CORAL & 0.9405  & 0.3464  & 0.5613  & 0.6914  & 0.7099  & 0.3528  & 0.4378  & 0.8215  & 0.7204  & 0.4779  & 0.3037  & 0.9370  & 0.6084  \\ \hline
        DANN & 0.9389  & 0.4372  & 0.6334  & 0.9107  & 0.9191  & 0.5752  & 0.6359  & 0.9002  & 0.8541  & 0.6855  & 0.4525  & 0.9386  & 0.7401  \\ \hline
        CDAN & 0.9481  & 0.5204  & 0.6365  & 0.9183  & 0.9237  & 0.5215  & 0.6329  & 0.9289  & 0.8817  & 0.6794  & 0.4801  & 0.9478  & 0.7516  \\ \hline
        SDAFDN & 0.9679  & 0.5216  & 0.8356  & 0.7540  & 0.8247  & 0.8169  & 0.8206  & 0.8530  & 0.7481  & 0.8104  & 0.5862  & 0.9561  & 0.7912 \\ \hline
        \textbf{Proposed method} & 0.9710  & 0.9198  &  0.6380 &  0.9213 & 0.9695  & 0.6196 & 0.9739 & 0.9455 & 0.9432 &  0.9557 & 0.6196 & 0.9662 & \textbf{0.8703}  \\ \hline
    \end{tabular}}
\end{table*}

\begin{table*}[ht] 
    \centering
    \caption{Transfer learning diagnosis results on PHM2009 dataset}
    \label{PHM}
    \resizebox{\textwidth}{!}{ 
    \begin{tabular}{|l|c|c|c|c|c|c|c|c|c|c|c|c|c|}
    \hline
        \textbf{Source Domain} & \multicolumn{2}{c|}{$\mathrm{M}_0\quad+\quad\mathrm{M}_1$} & \multicolumn{2}{c|}{$\mathrm{M}_2\quad+\quad\mathrm{M}_0$} & \multicolumn{2}{c|}{$\mathrm{M}_3\quad+\quad\mathrm{M}_0$} & \multicolumn{2}{c|}{$\mathrm{M}_1\quad+\quad\mathrm{M}_2$} & \multicolumn{2}{c|}{$\mathrm{M}_3\quad+\quad\mathrm{M}_1$} & \multicolumn{2}{c|}{$\mathrm{M}_3\quad+\quad\mathrm{M}_2$} &  \multirow{2}{*}{\textbf{Average}} \\ \cline{1-13}
        \textbf{Target Domain} & \multicolumn{2}{c|}{$\mathrm{M}_2 \qquad \quad \mathrm{M}_3$} & \multicolumn{2}{c|}{$\mathrm{M}_1 \qquad \quad \mathrm{M}_3$} & \multicolumn{2}{c|}{$\mathrm{M}_2 \qquad \quad \mathrm{M}_1$} & \multicolumn{2}{c|}{$\mathrm{M}_0 \qquad \quad \mathrm{M}_3$} & \multicolumn{2}{c|}{$\mathrm{M}_0 \qquad \quad \mathrm{M}_2$} & \multicolumn{2}{c|}{$\mathrm{M}_1 \qquad \quad \mathrm{M}_0$} &  \\ \hline
        JMMD & 0.5449  & 0.6058  & 0.6474  & 0.5128  & 0.7115  & 0.5641  & 0.6987  & 0.6827  & 0.5192  & 0.6955  & 0.6218  & 0.5096  & 0.6095  \\ \hline
        MK-MMD & 0.5160  & 0.6154  & 0.6506  & 0.4872  & 0.7436  & 0.5737  & 0.7019  & 0.7244  & 0.5096  & 0.6891  & 0.6442  & 0.6090  & 0.6221  \\ \hline
        CORAL & 0.4231  & 0.5128  & 0.5641  & 0.2212  & 0.6474  & 0.4936  & 0.5000  & 0.6250  & 0.3622  & 0.5449  & 0.4199  & 0.5417  & 0.4880  \\ \hline
        DANN & 0.5385  & 0.6122  & 0.6538  & 0.5064  & 0.7147  & 0.5769  & 0.6891  & 0.6923  & 0.4968  & 0.6827  & 0.6250  & 0.6603  & 0.6207  \\ \hline
        CDAN & 0.5513  & 0.5801  & 0.6571  & 0.5737  & 0.7212  & 0.5513  & 0.6763  & 0.7115  & 0.4936  & 0.6506  & 0.6378  & 0.5641  & 0.6141  \\ \hline
        SDAFDN & 0.5429  & 0.5821  & 0.6064  & 0.5077  & 0.5962  & 0.5058  & 0.4731  & 0.6538  & 0.4487  & 0.6436  & 0.5943  & 0.5135  & 0.5555 \\ \hline
        \textbf{Proposed method} & 0.6955  & 0.6827  & 0.7212  & 0.7596  & 0.7660  & 0.7051  & 0.7372  & 0.7788  & 0.7019  & 0.7468  & 0.6667  & 0.6058  & \textbf{0.7139}  \\ \hline
    \end{tabular}}
\end{table*}

\subsection{Ablation studies} 

\subsubsection{Comparison with multiple source and target domains}

In order to further validate the effectiveness of the proposed method, we constructed the following three scenarios on the basis of the PHM2009 dataset: baseline (CDAN), multi-source-single-target domain adaptation scenario (MSST), single-source-multi-target domain adaptation scenario (SSMT), and a direct combination of multi-target domains (c-MSMT). The detailed results are shown in Table \ref{ablationPHM}.
\begin{enumerate}
    \item[(a)] Baseline: The standard method of domain adaptation for single source and single target, CDAN \cite{long2018conditional}, was chosen as the baseline model for the comparison.
    \item[(b)] Comparison with MSST: As shown in Table \ref{ablationPHM}, in the multi-source-single-target scenario, each column represents the result of the transform from multiple source domains to a single target domain. For instance, $\mathrm{M}_0$/$\mathrm{M}_1$$\rightarrow$$\mathrm{M}_2$ represents the domain adaptation task from source domains $\mathrm{M}_0$ and $\mathrm{M}_1$ to target domain $\mathrm{M}_2$. Compared with the baseline, the increased number of source domains results in a better performance of MSST. However, the performance of MSST is inferior to the proposed method which is due to the fact that MSST provides no utilization of information from multiple target domains.
    \item[(c)] Comparison with SSMT: The single-source multi-target domain adaptation scenario represents the transform from a single source domain to multiple target domains. For example, the first group column in Table \ref{ablationPHM} represents the domain adaptation tasks of $\mathrm{M}_0$$\rightarrow$$\mathrm{M}_2$/$\mathrm{M}_3$ as well as $\mathrm{M}_1$$\rightarrow$$\mathrm{M}_2$/$\mathrm{M}_3$, and we report higher results for the case of different single source domains adapting to the same target domain. The results in Table \ref{ablationPHM} demonstrate that SSMT can well perform the domain adaptation task for multiple target domains and therefore achieves better results than the baseline. At the same time, SSMT lacks the labeled information of multiple source domains and therefore underperforms the proposed method.
    \item[(d)] Comparison with c-MSMT: In this scenario, we ignore the information between target domains and directly combine them into a single target domain, denoted by c-MSMT. By comparing c-MSMT with the proposed method, it can be found that the proposed method still significantly outperforms c-MSMT even when the source and target domains are the same. This may be attributed to the naive combination of multiple target domains performed by c-MSMT ignoring the information between target domains, which leads to the negative transfer phenomenon.
\end{enumerate}

\begin{table*}[htbp] 
    \centering
    \caption{Ablation study of the proposed method on PHM2009 dataset}
    \label{ablationPHM}
    \resizebox{\textwidth}{!}{ 
    \begin{tabular}{|l|c|c|c|c|c|c|c|c|c|c|c|c|c|}
    \hline
        \textbf{Source Domain} & \multicolumn{2}{c|}{$\mathrm{M}_0\quad+\quad\mathrm{M}_1$} & \multicolumn{2}{c|}{$\mathrm{M}_2\quad+\quad\mathrm{M}_0$} & \multicolumn{2}{c|}{$\mathrm{M}_3\quad+\quad\mathrm{M}_0$} & \multicolumn{2}{c|}{$\mathrm{M}_1\quad+\quad\mathrm{M}_2$} & \multicolumn{2}{c|}{$\mathrm{M}_3\quad+\quad\mathrm{M}_1$} & \multicolumn{2}{c|}{$\mathrm{M}_3\quad+\quad\mathrm{M}_2$} &  \multirow{2}{*}{\textbf{Average}} \\ \cline{1-13}
        \textbf{Target Domain} & \multicolumn{2}{c|}{$\mathrm{M}_2 \qquad \quad \mathrm{M}_3$} & \multicolumn{2}{c|}{$\mathrm{M}_1 \qquad \quad \mathrm{M}_3$} & \multicolumn{2}{c|}{$\mathrm{M}_2 \qquad \quad \mathrm{M}_1$} & \multicolumn{2}{c|}{$\mathrm{M}_0 \qquad \quad \mathrm{M}_3$} & \multicolumn{2}{c|}{$\mathrm{M}_0 \qquad \quad \mathrm{M}_2$} & \multicolumn{2}{c|}{$\mathrm{M}_1 \qquad \quad \mathrm{M}_0$} &  \\ \hline
        CDAN (Baseline) & 0.5513  & 0.5801  & 0.6571  & 0.5737  & 0.7212  & 0.5513  & 0.6763  & 0.7115  & 0.4936  & 0.6506  & 0.6378  & 0.5641  & 0.6141  \\ \hline
        MSST & 0.6731 & 0.6506 & 0.6987 & 0.7468 & 0.7692 & 0.6891 & 0.6378 & 0.7724 & 0.6859 & 0.7308 & 0.6603 & 0.5577 & 0.6894 \\ \hline
        SSMT & 0.6667 & 0.6474 & 0.6026 & 0.7340 & 0.7212 & 0.6186 & 0.6571  & 0.7244  & 0.6763  & 0.7436  & 0.6346  & 0.5577  & 0.6654 \\ \hline
        c-MSMT & \multicolumn{2}{c|}{0.6795} & \multicolumn{2}{c|}{0.6587} & \multicolumn{2}{c|}{0.6506} & \multicolumn{2}{c|}{0.6891} & \multicolumn{2}{c|}{0.7212} & \multicolumn{2}{c|}{0.6042} & 0.6672  \\ \hline
        \textbf{Proposed method} & 0.6955  & 0.6827  & 0.7212  & 0.7596  & 0.7660  & 0.7051  & 0.7372  & 0.7788  & 0.7019  & 0.7468  & 0.6667  & 0.6058  & \textbf{0.7139}  \\ \hline
    \end{tabular}}
\end{table*}


\begin{table*}[htbp] 
    \centering
    \caption{Comparison With Additional Source Domains on PHM 2009 dataset}
    \label{additionsource}
    \resizebox{\textwidth}{!}{ 
    \begin{tabular}{|l|c|c|c|c|c|c|c|c|c|c|c|c|c|}
    \hline
        \textbf{Source Domain} & \multicolumn{3}{c|}{$\mathrm{M}_0$\qquad$\mathrm{M}_1$\qquad$\mathrm{M}_2$}& \multicolumn{3}{c|}{$\mathrm{M}_0$\qquad$\mathrm{M}_1$\qquad$\mathrm{M}_3$} & \multicolumn{3}{c|}{$\mathrm{M}_0$\qquad$\mathrm{M}_2$\qquad$\mathrm{M}_3$} & \multicolumn{3}{c|}{$\mathrm{M}_3$\qquad$\mathrm{M}_2$\qquad$\mathrm{M}_1$} &  \multirow{2}{*}{\textbf{Average}} \\ \cline{1-13}
        \textbf{Target Domain} & \multicolumn{3}{c|}{$\mathrm{M}_3$}& \multicolumn{3}{c|}{$\mathrm{M}_2$} & \multicolumn{3}{c|}{$\mathrm{M}_1$} & \multicolumn{3}{c|}{$\mathrm{M}_0$} &  \\ \hline
        one2one & 0.5609  & 0.6218  & 0.7115  & 0.5513  & 0.6955  & 0.7212  & 0.5256  & 0.6506  & 0.6058  & 0.5032  & 0.5288  & 0.6314  & 0.6090 \\ \hline
        two2one & 0.6506 & 0.7468 & 0.7724 & 0.6731 & 0.7692 & 0.7308 & 0.6987 & 0.6891 & 0.6603 & 0.5577 & 0.6859 & 0.6378 & 0.6894 \\ \hline
        three2one & \multicolumn{3}{c|}{0.7853}&\multicolumn{3}{c|}{0.7821} & \multicolumn{3}{c|}{0.7212} & \multicolumn{3}{c|}{0.7340} & 0.7557 \\ \hline
    \end{tabular}}
\end{table*}


\begin{table*}[htbp] 
    \centering
    \caption{Comparison With Additional Target Domains on PHM2009 dataset}
    \label{additiontarget}
    \resizebox{\textwidth}{!}{ 
    \begin{tabular}{|l|c|c|c|c|c|c|c|c|c|c|c|c|c|}
    \hline
        \textbf{Target Domain} & \multicolumn{3}{c|}{$\mathrm{M}_3$}& \multicolumn{3}{c|}{$\mathrm{M}_2$} & \multicolumn{3}{c|}{$\mathrm{M}_1$} & \multicolumn{3}{c|}{$\mathrm{M}_0$} & \multirow{2}{*}{\textbf{Average}} \\ \cline{1-13}
        \textbf{Source Domain} & \multicolumn{3}{c|}{$\mathrm{M}_0$\qquad$\mathrm{M}_1$\qquad$\mathrm{M}_2$}& \multicolumn{3}{c|}{$\mathrm{M}_0$\qquad$\mathrm{M}_1$\qquad$\mathrm{M}_3$} & \multicolumn{3}{c|}{$\mathrm{M}_0$\qquad$\mathrm{M}_2$\qquad$\mathrm{M}_3$} & \multicolumn{3}{c|}{$\mathrm{M}_3$\qquad$\mathrm{M}_2$\qquad$\mathrm{M}_1$} &  \\ \hline
        one2one & 0.5032  & 0.6058  & 0.7212  & 0.5288 & 0.6506  & 0.7115  & 0.6314  & 0.6955  & 0.6218  & 0.5256  & 0.5513  & 0.5609  & 0.6090 \\ \hline
        one2two & 0.4679  & 0.6186  & 0.7628  & 0.5865  & 0.6346  & 0.7340  & 0.6763  & 0.7083  & 0.6474  & 0.5481  & 0.5192  & 0.5321  & 0.6197 \\ \hline
        one2three & 0.4840  & 0.5833  & 0.7115  & 0.5353  & 0.6442  & 0.7532  & 0.5994  & 0.5962  & 0.6122  & 0.5609  & 0.5096  & 0.5417  & 0.5943 \\ \hline
    \end{tabular}}
\end{table*}

\begin{figure*}[htbp]
\centering
\subfloat[]{\includegraphics[width=1.35in]{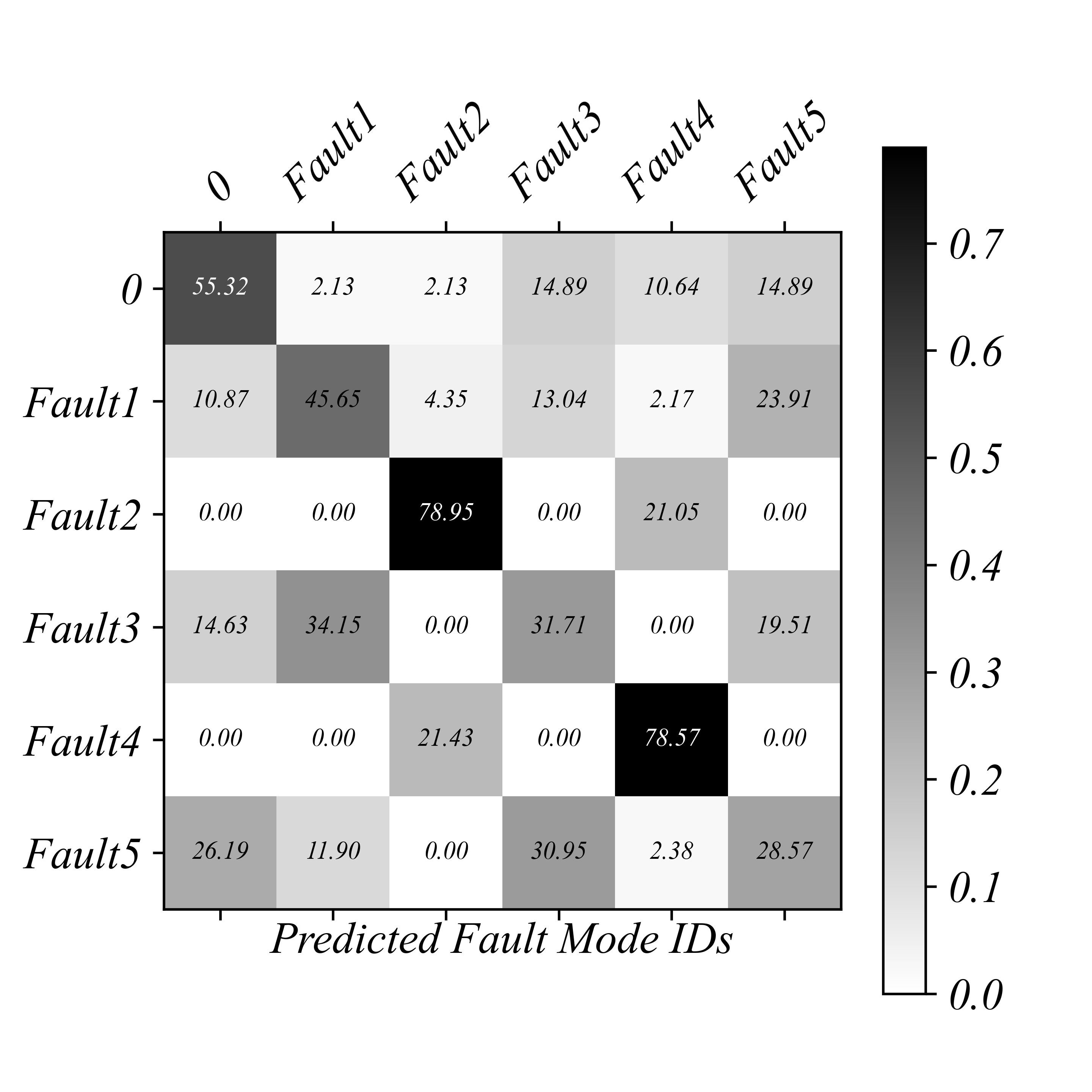}%
\label{confuPHMJMMD}}
\hfil
\subfloat[]{\includegraphics[width=1.35in]{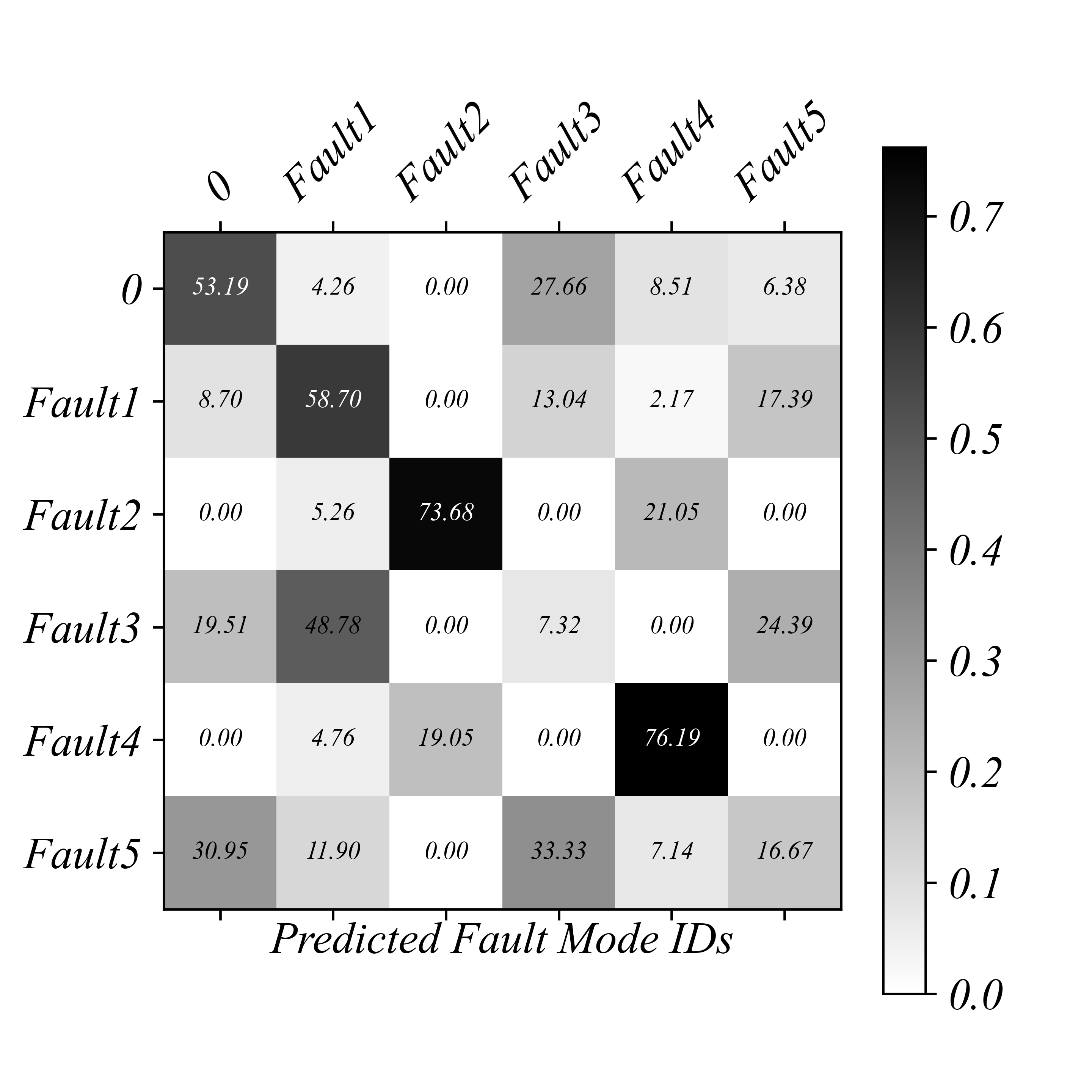}%
\label{confuPHMMKMMD}}
\hfil
\subfloat[]{\includegraphics[width=1.35in]{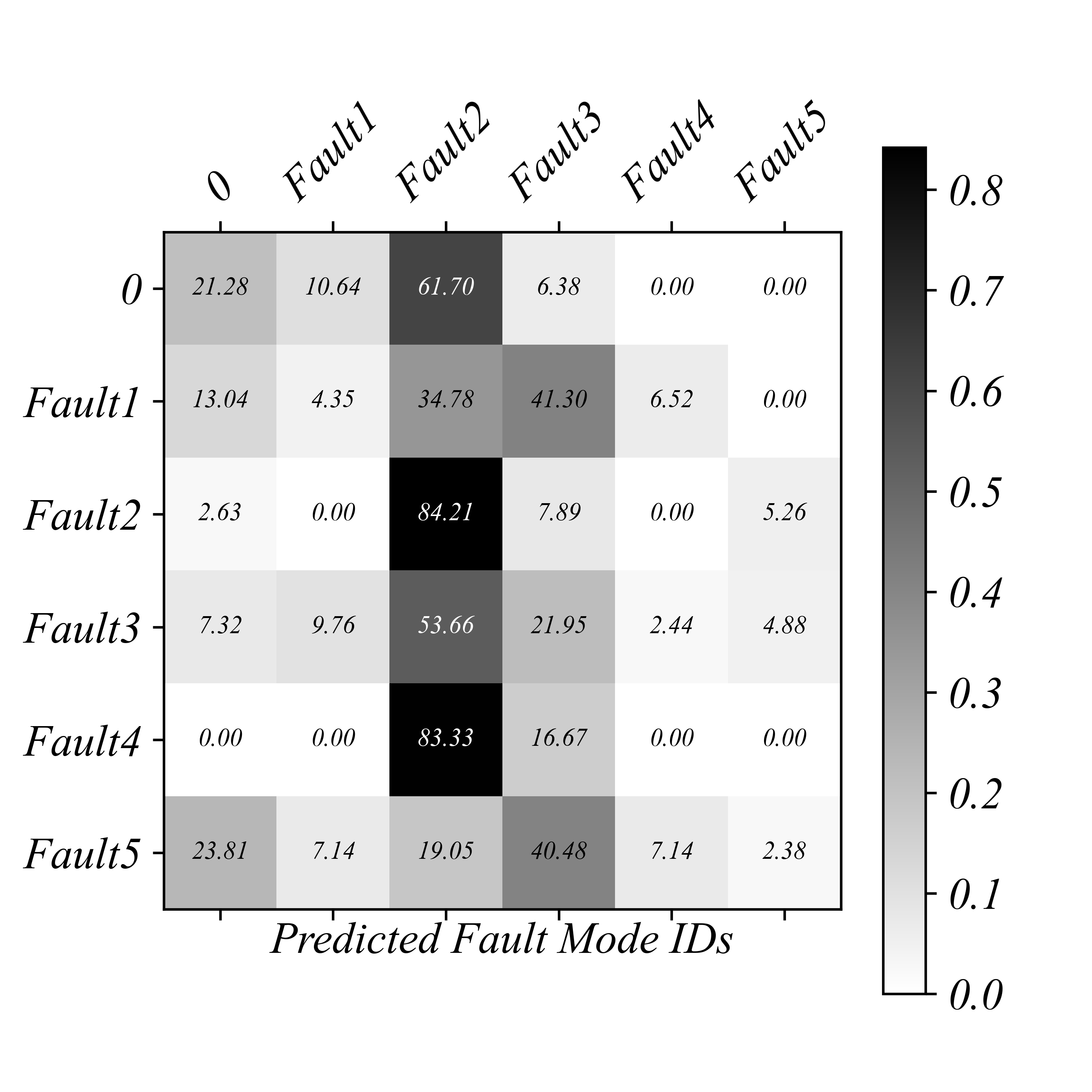}%
\label{confuPHMCORAL}}
\hfil
\\
\vspace{-1.0em} 
\subfloat[]{\includegraphics[width=1.35in]{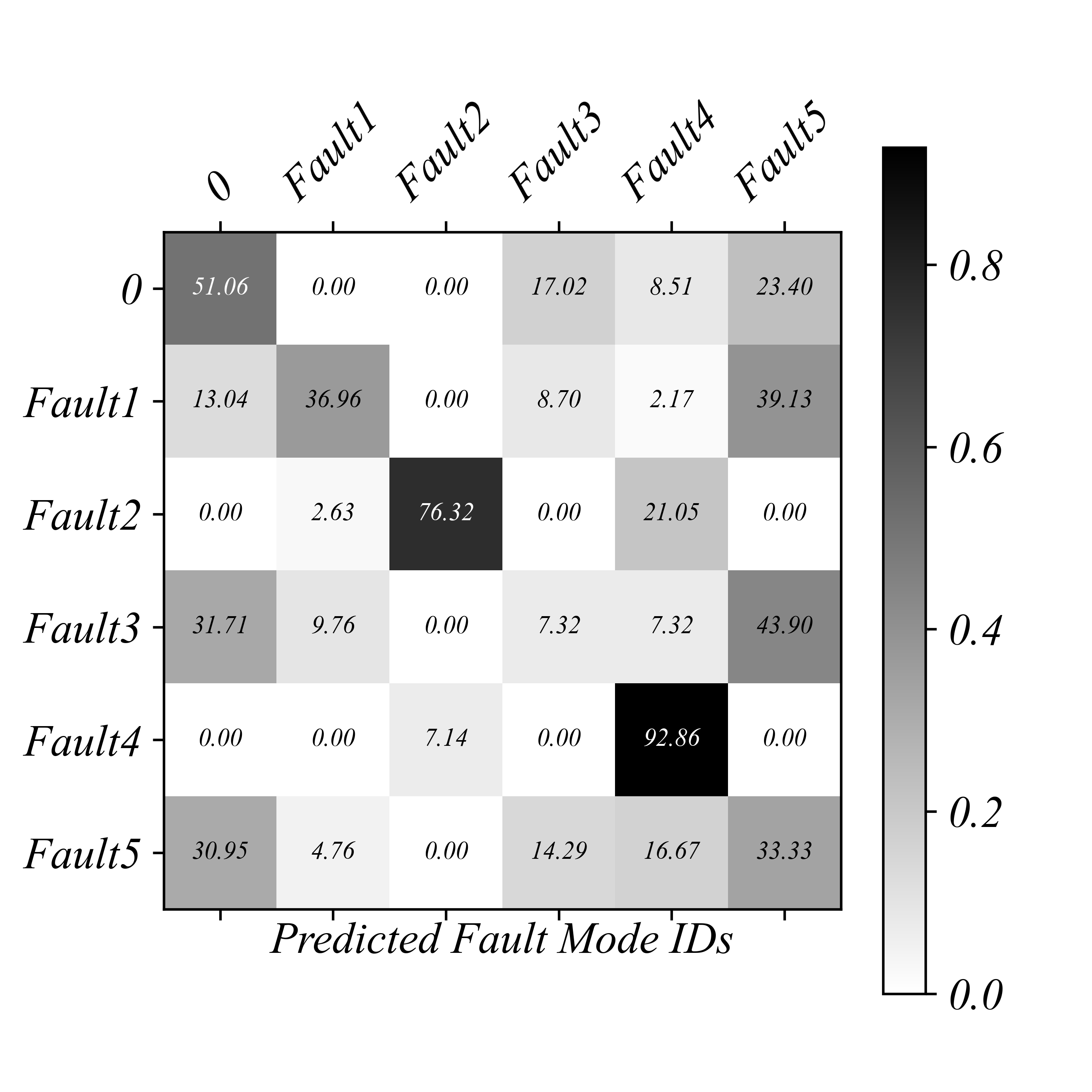}%
\label{confuPHMDANN}}
\hfil
\subfloat[]{\includegraphics[width=1.35in]{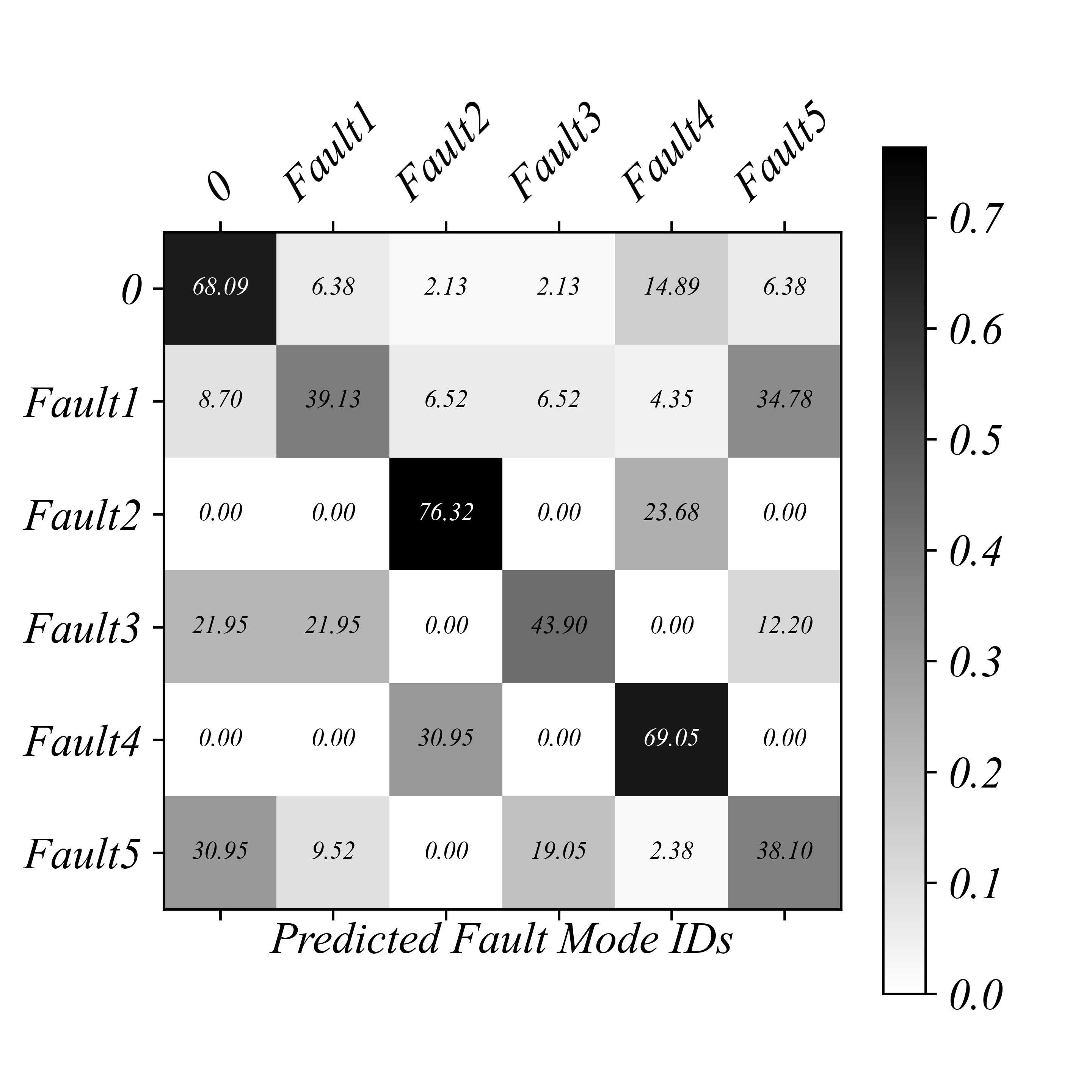}%
\label{confuPHMCDAN}}
\hfil
\subfloat[]{\includegraphics[width=1.35in]{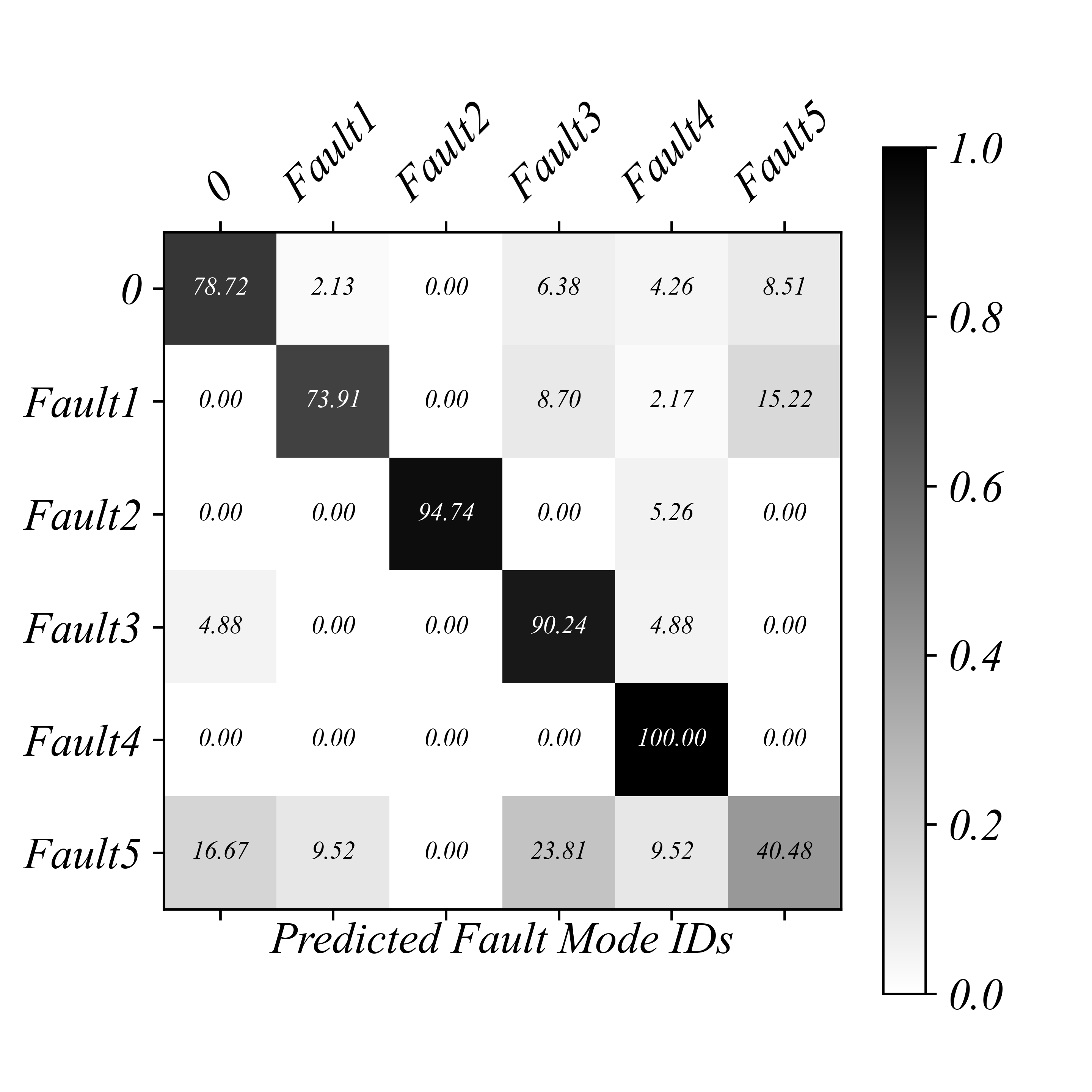}%
\label{confuPHMMSMT}}
\hfil
\caption{Confusion matrices obtained from domain adaptation task $\mathrm{M}_2$/$\mathrm{M}_0$ $\rightarrow$ $\mathrm{M}_1$/$\mathrm{M}_3$ for target domain $\mathrm{M}_3$ on PHM2009 dataset.  (a) JMMD. (b) MK-MMD. (c) CORAL. (d) DANN. (e) CDAN. (f) WJMMD-MDA (proposed method).}
\label{confusion}
\end{figure*}

\subsubsection{Comparison with additional source and target domains}

To explore the effect of the number of source and target domains on the performance of the proposed method, we designed the following experiments based on the PHM2009 dataset. In detail, the multi-source domain scenario includes 'one2one', 'two2one', and 'three2one', while the multi-target domain scenario includes 'one2one', 'one2two', and 'one2three'. The experimental results are shown in Table \ref{additionsource} and Table \ref{additiontarget}. The results presented in Table \ref{additionsource} demonstrate that the proposed method possesses better performance in domain adaptation tasks as more source domains are involved. This is facilitated with the introduction of more labeled information from additional source domains. Moreover, the results in Table \ref{additiontarget} indicate that more target domains make it even more challenging for the proposed method to perform domain adaptation effectively. This may be mainly caused by additional target domains introducing more unlabeled data, thus complicating the data distribution. 

\subsection{Visualization Analysis} 
The confusion matrices obtained from the proposed method as well as the comparison methods for domain adaptation fault diagnosis on PHM2009 dataset are shown in Fig. \ref{confusion}. 
Fig. \ref{confuPHMJMMD} to Fig. \ref{confuPHMCDAN} show that significant confusion exists between the various classes of the results obtained from domain adaptation by the comparison methods.
Specifically, almost all of the compared domain adaptation methods fail to accurately diagnose Fault 1, Fault 3, and Fault 5, while the proposed method solves this problem to a large extent. This is due to the fact that such single-source-single-target domain adaptation methods are incapable of obtaining effective information from multiple source domains and multiple target domains.
On the contrary, benefiting from the ability to utilize information from multiple source domains and multiple target domains, which is innovatively introduced in this study, the result obtained by the proposed method, illustrated in Fig. \ref{confuPHMMSMT}, represents the optimal performance.

\section{Conclusion}
In this paper, for the first time in the field of fault diagnosis, we propose a multi-source-multi-target unsupervised domain adaptation method, which can effectively utilize the labeled information from multiple source domains and perform the domain adaptation task on multiple target domains simultaneously. The core idea of this paper is to compute the JMMD distances between the data of multiple source and target domains and assign weights to them, thus reducing the distributional differences between the data of source and target domains. We validate the effectiveness of the proposed method through comprehensive comparative experiments on several datasets. The experimental results demonstrate that the proposed method substantially outperforms the compared methods in the tasks of unsupervised domain adaptation fault diagnosis.\cite{qiaoSurveyWindTurbine2015a}

\section*{Acknowledgment}

This work was supported by the National Natural Science Foundation of China (No.62276230).

\bibliographystyle{unsrtnat}
\bibliography{ref.bib}

\newpage

\end{document}